# Real-life Implementation of Internet of Robotic Things Using 5 DoF Heterogeneous Robotic Arm

Sayed Erfan Arefin [1], Tasnia Ashrafi Heya[1], Jia Uddin[1]
Department of Computer Science and Engineering,
BRAC University, 66 Mohakhali, Dhaka
{erfanjordison,tasnia.heya,tajwar.hoque}@gmail.com, jia.uddin@bracu.ac.bd

**Abstract.** Establishing a communication bridge by transferring data driven from different embedded sensors via internet or reconcilable network protocols between enormous number of distinctively addressable objects or "things", is known as the Internet of Things (IoT). IoT can be amalgamated with multitudinous objects such as thermostats, cars, lights, refrigerators, and many more appliances which will be able to build a connection via internet. Where objects of our diurnal life can establish a network connection and get smarter with IoT, robotics can be another aspect which will get beneficial to be brought under the concept of IoT and is able to add a new perception in robotics having "Mechanical Smart Intelligence" which is generally called "Internet of Robotic Things" (IoRT). A robotic arm is a part of robotics where it is usually a programmable mechanical arm which has human arm like functionalities. In this paper, IoRT will be represented by a 5 DoF (degree of freedoms) Robotic Arm which will be able to communicate as an IoRT device, controlled with heterogeneous devices using IoT and "Cloud Robotics".

## 1 Introduction

Robotics is a field of study in which reprogrammable, multifunctional controller intended to move material, parts, instruments, or concentrated gadgets through different customized and programmed movements or motions to execute divergent tasks. It includes the concept of robotic arms and the activities of this is very analogous to the activities of human arms. Automated arms are the crucial piece of the considerable number of ventures where they perform different diverse duties, for example, welding, trimming, picking and putting and so forth, in addition to that, the greatest favorable position of these arms is that, it can work in high-risk regions and furthermore in the regions which cannot be gotten to by human [1]. Networked or pre-programmed robotic arms have been exceptionally fruitful for industries because of having high endurance, speed, and accuracy in structured factory environments. Though it has made incredible progress in modern factorial applications, transportation frameworks, and security applications, but the development of networked robotics has severe limitations in case of resource, information, and communication inalienable in the current infrastructure as, all calculations are done on the locally available robots, which have constrained computing abilities and accessing information is likewise confined to the aggregate storage of the system. Cloud Robotics is a concept to get rid of these constraints which results in more efficient, intelligent but cheap robotic solutions. [2]

Cloud robotics concept is a compound result of cloud technologies and robotics where the lacking of robotic arms such as computational capacity, memory and storage can be demolished by the empowerment of offload computation-intensive and storage-intensive jobs into the cloud [3]. If robotic arms follows this concept, there can be four possible advantages which are, a) remote libraries of images, maps, trajectories, and object data can be easily accessible, b) accessibility of statistical analysis, learning, and motion planning with parallel grid computing, c) trajectories, control policies, and outcomes can be shared by the arms by Collective Robot Learning and last but not the least, d) using crowd sourcing with the help of analyzing images, classification, learning, and error recovery by human expertise which can be called Human Computation in short [4]. Another interesting absorption of IoT based robotic arms using cloud robotics can be the increasing ability to learn the techniques of tasks done by human in a very quick manner which is called "Rapid Learning". For example, Fanuc, Japanese industrial robots which are capable of figuring out the best way to perform a particular task in around eight hours with almost 90% accuracy by observing how humans are performing the task through brute forcing deep reinforcement learning with continual trial and error [5]. Surveying over 300 manufacturers, it was found that, though media has been conveying frightening ideas about cloud robotics in manufacturing process but most of the surveyed have shared that there are no such problems, in fact, a large amount among them have expressed that cloud robotics is the major reason behind the security improvement [6].

In this paper, it will be elaborated that how an important part of robotics which is, robotic arm, works as an IoRT device and what are the scopes and advantages of the application of cloud robotic and IoRT in robotic arms, by implementing a 5 DoF heterogenic robotic arm based on IoRT. Cloud Robotics is the most vital part of IoRT where robotic arms will act as IoT devices by being connected with Internet and will communicate using the concept of cloud robotics. First attraction of using IoT services is, robotic arms can get smarter and this will be very beneficial.

## 2   Background Research

### 2.1   Internet of Things (IoT) and Internet of Robotic Things (IoRT)

Transferring data over network via Internet without any human-to-human connections or human-to-machine relations is the benefit of IoT which integrate components, for example, mobility support, divergent geo-distribution, wireless accessibility, and a higher-level stage for a particular number of IoT services [8].

Moreover, Real-time control and analytics by data transmission ensuring fast response time, are ensured by IoT applications. IoT devices can communicate within themselves without interventions. They are provided with a kind of URI (Uniform Resource Identifier) named URN (Uniform Resource Name) in order to distinguish each IoT service nodes individually with name or id [9].

IoRT concept is using the IoT field in the robotics area so that robots can act and perform in a better and smarter way. This concept is theoretically explained in [10] briefly and described five layers of the infrastructure. Moreover, they have provided the guideline and overview about the probable protocols which can be followed in each layer to build an IoRT device or robots or robotic arms which has been shown in Fig. 1.

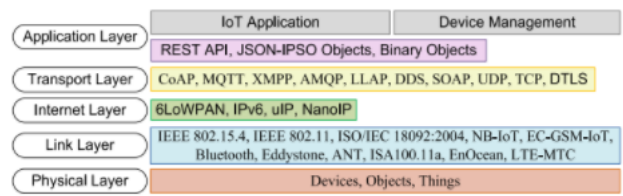

**Fig. 1** IoRT Infrastructure with possible protocols [10].

According to [10], the main difference between IoT and IoRT is that, the robots working as IoT devices not only can communicate but also can take real actions in the physical world. Different service providers establish the connectivity between the robots and the data management system or cloud. Moreover, full support for deployment, integration, maintenance as well as consulting services is included and thus, service requires the highest rate of growth as well. In order to perform other connectivity actions by the support of programming along with hardware, software is a vital part which is needed to be incorporated with the robots.

### 2.2 Cloud Robotics

Cloud robotics is the concept of network-connected robots which use the benefits of parallel computation and data sharing by connecting with the Internet using the indefinite resources of big data and performing collaboration as well as rapid collective learning [8]. Cloud computing provides solution to overcome the limitations and extend the capacities of networked robotics. Cloud robotics is the amalgamation of two complementary clouds which are, ad-hoc cloud where participating robots perform machine-to-machine (M2M) communications and an infrastructure cloud which follows the machine-to-cloud (M2C) communications [2].

**Ad-hoc Cloud.** There are three benefits of ad-hoc cloud. At first, a virtual ad-hoc infrastructure can be formed by pooling the capacity of computation of individual robot together. Next, collaborative decisions can be chosen by exchanging knowledge among the unit of collaborative computation, in different applications of robotics.

Finally, it helps the robots which are out of range, to send request to the storage system to establish communication. [2]

**Infrastructure Cloud.** A usable format for robots of arranged data about the situations around are given and cloud enabled robots are able to learn from the provided libraries combined with skills records according to hazardous situations or tasks. [2]

### 2.3 Robotic Arm

Robotic arms represents basic supplementary machines utilized for the extensive range of operations, for example, control such as transport of raw materials, or generation such as painting, welding or gathering and they symbolizes kinematics with magnificent portability and adeptness and appropriate work space scope. Various applications of robotic arms [7] play vital roles in different aspects and some of them are,

**Industrial applications.** robotic arms are used in different welding purpose such as Arc Welding, Electron Beam, Flux Cored Welding, Laser Welding, Mig Welding, Plasma Cutting, Plasma Welding, Resistance Welding, Spot Welding, Tig Welding, and Welding Automation.

**Material handling applications.** Collaborative, Dispensing, Injection Molding, Machine Loading, Machine Tending, Material Handling, Order Picking, Packaging, Palletizing, Part Transfer, Pick and Place, Press Tending and Vision.

**Other applications.** 3D Laser Vision, Appliance Automation, Assembly, Bonding / Sealing, Cleanroom, Coating, Cutting, Drilling, Fiberglass Cutting, Foundry, Grinding, Material Removal, Meat Processing Automation, Milling, Painting Automation, Polishing, Refueling, Routing, Spindling, Thermal Spray and Water jet.

Though there are several useful applications of robotic arms but there exists some limitations as well. For instance, lacking opportunities of remotely accessibility in hazardous areas where human cannot operate them directly because of, lack of communicating skills of the arms on those places, lacking of quick learning skills from both human and other intelligent robots to become prepared for any critical situation. These limitations can be solved by using the IoRT services which can be very beneficial for increasing scopes of robotic arms and their rapid learning skills.

### 3 System Implementation

The system consists of 5 layers following the model of IoRT architecture. The layers are: the hardware layer, the network layer, the internet layer, the infrastructure layer and the application layer [10]. In what manner all these layers are used in this 5 DoF Heterogenic Robotic Arm, are described below.

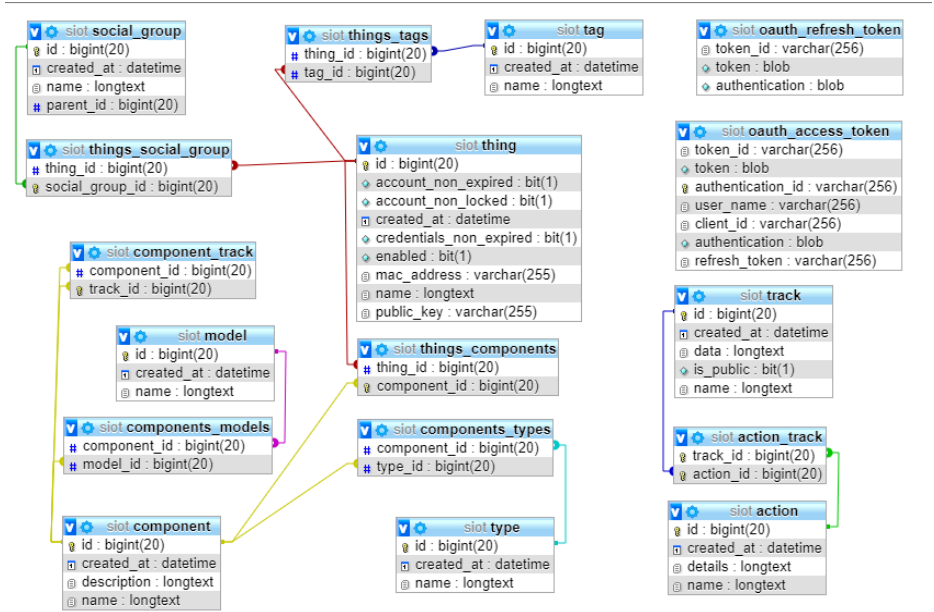

**Fig. 2** Schema Diagram of the Robotic Arm.

### 3.1 Hardware Layer

The main robotic arm and the controller of our implementation of the IoRT model is comprised of the following:

**Raspberry Pi 3.** The Raspberry Pi 3 is a micro size computer capable of controlling at least 6 servo motors. The Pi has a Ethernet Port. We will be using that to connect with the internet. [13]
.

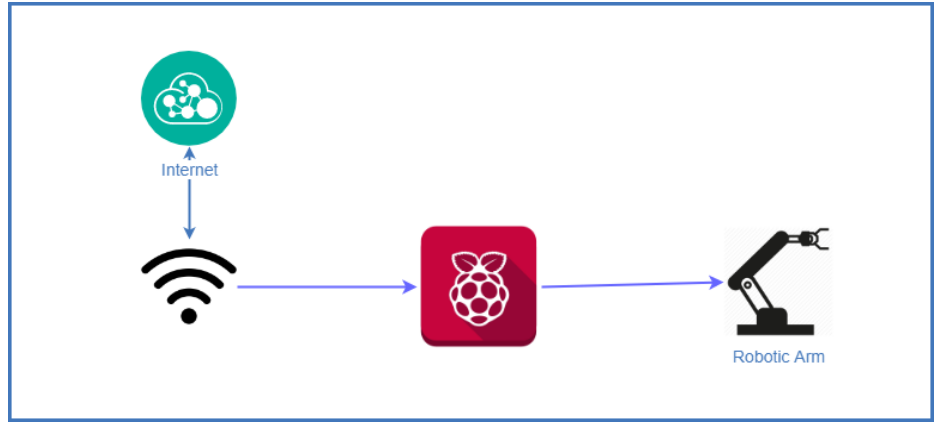

**Fig. 3** Network layer and Hardware layer.

**Robotic arm.** A 5 DOF robotic arm was developed for the implementation which consists of Wrist, Arm, Shoulder and Base. The movements of the arm is shown in the Fig. 3.

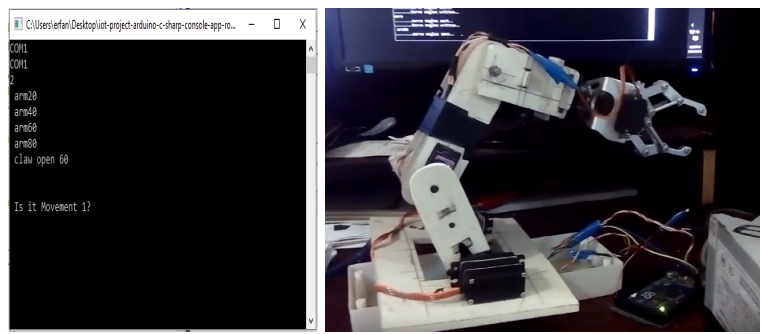

**Fig. 4** Movement of the Arm.

*Servo Motor.* We used MG995 servo motors for the movements of the joints, which has metal gear dual ball bearing with the following specifications: 120 degrees (60 in each direction) of movements, 55 gram of weight, stall torque of 8.5 kgf•cm (4.8 V ), 10 kgf.cm (6 V) [15].

*Main body construction.* Two servos were used in the shoulder movements in order to provide more torque. The length of the wrist, arm and shoulder were 4.5 cm, 10cm and 6.5 cm consequently. The material used to make the main body was 3mm plastic wood and we used screws and hot glue to attach the servo motors and the main structure.

*End Effector.* The end effector is the part of a robotic arm that interacts with the environment physically. The most suitable end effector for our purpose will be a general-purpose grabber. This is a MKII robotic claw developed by DAGU which has 2 inches of open-close range of the fingers [16]. For the end effector operation and the wrist horizontal rotation we used two different types of servos. These are metal gear (micro size) servos with an operating voltage of 6 volts and delivered 44.4 oz-in. the highest torque at 0.18 sec/60°. The weight is 20g and has a rotation of 180 degrees.

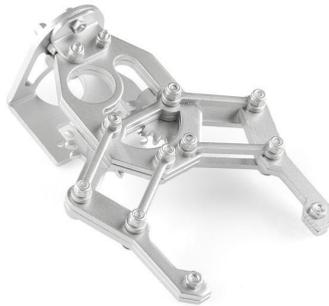

**Fig. 5** End Effector

*Power supply.* We used a 450W P5 personal computer power supply as the power supply for the robotic arm. By shorting pin 14 and 15 we were able to turn the power supply on. We used the 12V and Ground through a power distribution board to lower the voltage to 6V in order to operate the robotic arm properly.

### 3.2 Network Layer

In order to communicate with the internet, we are using a Raspberry Pi to connect to the internet. [14]. Using a conventional Ethernet cable (CAT5 or CAT6 with RJ45 connectors). By operating with IEEE 802.11 via a RJ45 connector, the connection to the internet and the hardware layer is built.

### 3.3 Internet Layer

Internet connection plays a vital role in communication of IoRT infrastructure. Communication protocols of IoT is a part of this layer to increase energy and resource efficiency as well as light weight data processing in robotic arms [10]. We are using the TCP/IP protocol to establish connection and using several cloud modules and services to make the system workable.

## 3.4 Infrastructure Layer

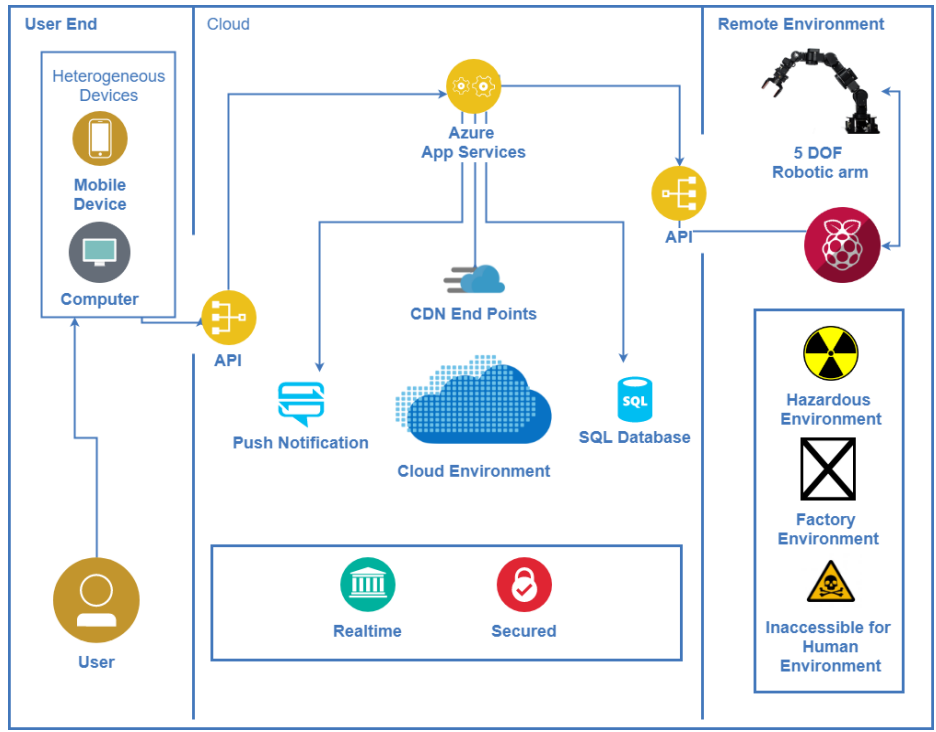

**Fig. 6** Infrastructure.

This layer can be considered as the most significant one because of the robotic service centric cloud storage along with middleware, business process, and big data, built based on IoT. This layer is agglomerated with 5 diversely yet related structure (robotic cloud platform, M2M2A cloud platform support, IoT business cloud services, Big Data services, and IoT cloud robotics infrastructure) [12]. Machine or the robotic arm will be receiving and giving feedback via the cloud. In this layer the cloud robotic logics are molded in. For our purpose to implement the IoRT structure, we have used Microsoft Azure and its services. The following services are used for our implementation. For the cloud section of the Internet of Robotic Things Model, we have used Microsoft Azure to implement our system modules and logics and Azure App Services to implement our primary logics. Additionally, we have used Azure SQL, Push Notification Service, Service Hub and the Content Delivery Network (CDN).

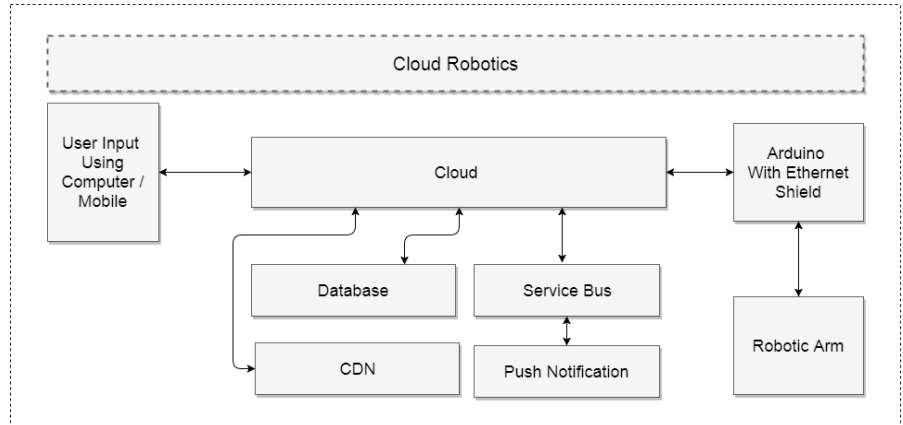

**Fig. 7** Cloud Robotics Representation.

**Azure App Services.** Azure app services receive requests and conveys the command to the robot through API services, which only communicates with the robot with Json. Additionally, this also stores the command patterns in the database. With a new node every time. If a previously recorded pattern is found then, it is prompt to the user that if this is the command she is trying to use. This functionality is handled by the application layer. But app services manage all the requests from the application layer to work with CDN, Service bus, Push notification and Mongo DB.

**Service Bus.** Service bus is used to support push notification and to receive the requests from the user and robot by creating separate queues which will reduce the loss of commands.

**Push Notification.** Push notification is used to send push messages both to user and robot to have a real time communication.

**Mongo DB.** Mongo db is used as the database. The root node divides into two sub nodes, one is "On going" and another one is "Learning". Each sequence of command is stored in the "On going" node with the timestamp. If any sequence is identified as successful and commonly used by the App Service, it will store that node to the "Learning" node.

**CDN.** CDN (Content Delivery Network) is used to have instances of the App service and Mongo DB to every regional node so that the user and robotic hand both can have faster access.

### 3.5 Application Layer

The application layer consists of three sections. Firstly, the Ardunio part, which receives the angles to be set from the infrastructure layer and after the operation it sends an acknowledgement to the infrastructure. Secondly, the App services of the Infrastructure layer plays the most vital part. Using the Azure sdks with php the infrastructure layer App services app was developed, that received and sent commands and notifications via service bus. Another part of this application was to recognize patterns of commonly used commands sequence and store them in the "learning" node of Mongo db. Again, every time a sequence of commands are received in the service bus, the application searches for the received pattern in the "learning" node to reuse and prompt to the user. If user accepts it, then the pattern will be reused. We implemented quick learning of IoRT by this. Thirdly, a user end console-based software was developed using C# and .NET that communicated with the Infrastructure layer.

## 4  Benefits of 5 DoF IoRT Based Robotic Arm

**Heterogeneity.** The robotic arm can be controlled from any kind of device having an internet connection because the infrastructure layer receives the commands via api and Azure app services. Depending on the devices a platform dependent app may be required.

**Remotely Controllable.** This robotic arm can be controlled remotely from any place via Internet as it takes instructions from cloud storage by Internet. Therefore, it is easy to monitor and control.

**Smart Decision Taking.** Smart robotic arms are able to take decisions in different situation by learning human activities or from other robotic activities which can be very helpful in different situation of different tasks.

**Communication.** Communication process is more efficient and easier. This arm can communicate about all the difficulties or it can request for support from any situation or range using cloud service. More efficient communication results in an increase of productivity, safety and quality of product.

**Collective learning ability.** This robotic arm uses the services of cloud robotic as well as IoRT which provides the ability of learning from both human and other robotic arms. Thus, it can take smart decisions based on those learning and can perform needed actions according to that.

**Performing difficult tasks.** Smart robotic arms can handle difficult tasks with more accuracy than human such as lifting and moving heavy objects.

## 5  Scopes

Robotic arms are already being used for,

**Clinical Services.** It will not be very optimistic if it is predicted that in near future robotic arms will perform complex operations as medical robotics is a huge field and rapidly developing and smart robotic arms are able to perform minor operations because of their collaborating learning capacities.

**Working as Third Hand.** In difficult jobs which human are not able to perform alone with two hands, this robotic arm can perform as a third hand in those jobs where robotic arm will hold the object while working on it.

**Retrieving Suspicious Objects.** In case of dangerous situations where sending human is a risk, for example, in the situation where a bomb needs to be disposed, this smart robotic arm can solve handle the situation by rapid learning instead of risking a human life.

**Surveillance Activities.** In case of surveillance activities this smart robotic arms can be very useful and efficient in performance.

**Sorting Colorful Objects.** Smart robotic arms can be used for sorting objects by color as they are capable of decision taking by collective learning. They can arrange things according to color preference in needed tasks such as in clothing or pencil factories, this can be a very useful activity while packaging.

# 6 Conclusion

In this paper, we tried to implement and improvise the concept of Internet of Robotic Things (IoRT) with a 5 DoF heterogeneous robotic arm which will be a great addition in the robotic field as it will work as a smart IoRT device connected with Internet and able to communicate efficiently regarding desired tasks, in addition, it can take decisions by rapid collective and collaborative learning from human activities as well as activities from other robotic arms. This smart robotic arm can be beneficial in various fields such as clinical services, security services, industrial services, differentiating objects with colors or patterns and it can perform tasks in dangerous situations instead of endangering human lives.